\documentclass{article}

\usepackage[preprint]{neurips_2024}

\usepackage[utf8]{inputenc} % allow utf-8 input
\usepackage[T1]{fontenc}    % use 8-bit T1 fonts
\usepackage{hyperref}       % hyperlinks
\usepackage{url}            % simple URL typesetting
\usepackage{booktabs}       % professional-quality tables
\usepackage{amsfonts}       % blackboard math symbols
\usepackage{nicefrac}       % compact symbols for 1/2, etc.
\usepackage{microtype}      % microtypography
\usepackage{xcolor}         % colors
\usepackage{microtype}

\usepackage{graphicx}

\usepackage{subfigure}
\usepackage{booktabs} % for professional tables
\usepackage{float}
\usepackage{subcaption} 
\usepackage{xcolor,colortbl}
\definecolor{forestgreen(web)}{rgb}{0.13, 0.55, 0.13}
\usepackage{listings}

\usepackage{hyperref}
\usepackage{amsmath}
\usepackage{amssymb}
\usepackage{mathtools}
\usepackage{amsthm}
\usepackage{wrapfig}
\usepackage[capitalize,noabbrev]{cleveref}
\usepackage{tikz}
\usetikzlibrary{matrix}
\usepackage{bbm}
\setcitestyle{square, comma, numbers}
\theoremstyle{plain}
\newtheorem{theorem}{Theorem}[section]

\theoremstyle{definition}
\newtheorem{definition}[theorem]{Definition}

\theoremstyle{remark}

\title{Towards Narrowing the Generalization Gap \\ in Deep Boolean Networks}

\author{%
 Youngsung Kim\\
  Department of Artificial Intelligence; Department of Electrical and Computer Engineering\\
  Inha University\\
  \{\texttt{yskim.ee@gmail.com}\} \\
}

\begin{document}

\maketitle

\begin{abstract}

    The rapid growth of the size and complexity in deep neural networks has sharply increased computational demands, challenging their efficient deployment in real-world scenarios. Boolean networks, constructed with logic gates, offer a hardware-friendly alternative that could enable more efficient implementation. However, their ability to match the performance of traditional networks has remained uncertain. This paper explores strategies to enhance deep Boolean networks with the aim of surpassing their traditional counterparts. We propose novel methods, including logical skip connections and spatiality preserving sampling, and validate them on vision tasks using widely adopted datasets, demonstrating significant improvement over existing approaches. Our analysis shows how deep Boolean networks can maintain high performance while minimizing computational costs through 1-bit logic operations. These findings suggest that Boolean networks are a promising direction for efficient, high-performance deep learning models, with significant potential for advancing hardware-accelerated AI applications.

%This work bridges the gap between theoretical efficiency and practical performance in hardware-friendly neural architectures.

% In this paper, we provide empirical evidence advocating the development of deep boolean networks as a key strategy for efficient neural networks at scale. Our approach involves the direct optimization of networks at the logic operation level, aiming to simplify the design loop from the deployment of machine learning algorithms to the AI hardware chip. Through exploring the integration of recent deep learning techniques into vanilla boolean (logic) networks, we encourage AI researchers to channel their efforts into the development of networks that contribute to \textcolor{forestgreen(web)}{Green}AI.
\end{abstract}

\section{Introduction}
\label{sec:introduction}

 In machine learning and artificial intelligence, the drive for high performance has led to the development of larger, more computationally intensive models. The training and inference costs for these models have increased dramatically, as exemplified by the increase in computational demands from AlexNet in 2012 to AlphaGo Zero in 2017, which rose by up to 300,000 times \citep{schwartz2020green}. These larger networks often demonstrate enhanced performance and generalization in complex scenarios \citep{BengioLecun2007_scaling, merkh2019stochastic}. This reflects the principle that achieving superior results typically requires more substantial computational resources, particularly in addressing complex AI challenges.

 However, large models that do not consider hardware implementation raise concerns about the additional optimization costs and energy consumption required for deployment.
 To address the growing demand for computational efficiency, researchers have explored various strategies, including pruning \citep{Zheng2022prunning}, compression \citep{wang2022compression, wang2022deep}, and low-bit quantization \citep{liu2022robust, kuzmin2022fp8}. Additionally, there have been efforts to develop energy-efficient hardware \citep{natelec2023, AMEET2020spectrum, sarah2023spectrum}.

\begin{figure}[t]
    \centering    \includegraphics[width=0.90\textwidth]{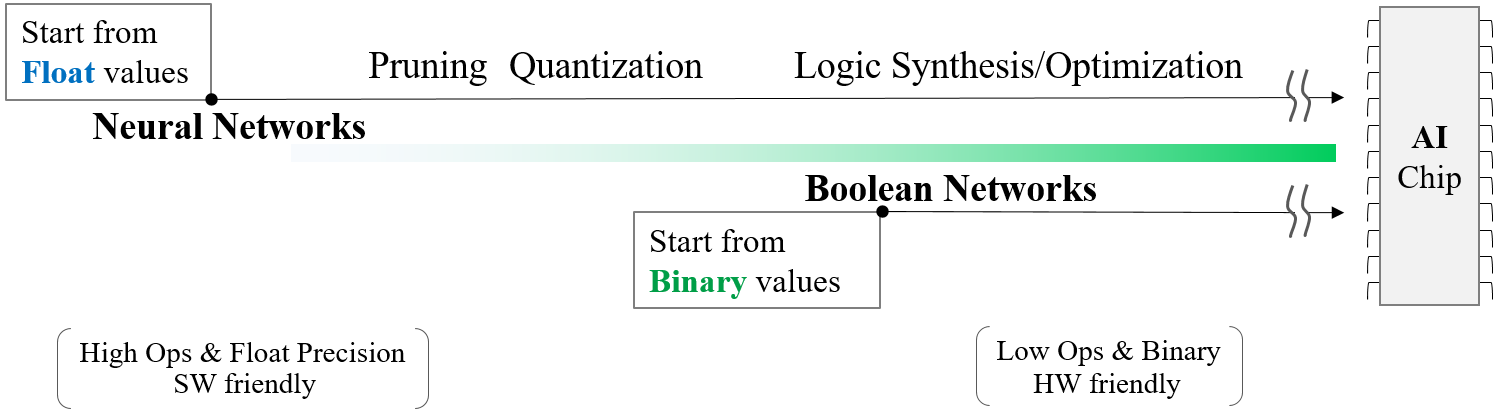}
    \vskip -0.05in
    \caption{In existing literature, optimizing deep neural networks often requires techniques like pruning and quantization when using floating-point precision. In contrast, boolean logic networks operate directly with binary expressions, potentially simplifying hardware implementation without the need for additional methods. This aligns with the principles of Hardware-friendly AI, where logic minimization/optimization can be conducted concurrently, including technology-independent or technology-dependent logic design and optimization based on specific logic libraries.
    }\label{fig:hwfriendly}
    \vskip -0.12in
\end{figure}

 Given these challenges, we promote a paradigm shift: instead of optimizing neural networks for existing hardware with additional optimization after finalizing the network architecture and training, we advocate for directly designing and training optimal chip components, specifically logic gates, to minimize the need for further optimization. This approach, illustrated in Figure~\ref{fig:hwfriendly}, leverages the natural alignment of Boolean functions with digital chip hardware, which predominantly operates using logic gates like \texttt{NAND} and \texttt{NOR}. The foundations of this approach are rooted in early AI research on logic networks, such as the Logic Theorist and General Problem Solver \citep{newton1934principia, newell1956logic, newell1957programming, newell1959report}. 
 %One example to describe their efficiency is that, the XOR problem can be efficiently solved using Boolean logic, specifically the \texttt{XOR} operation $ (A + B - 2 AB )$ for two Boolean inputs $(A, B)$ \citep{Rosenblatt1957}.

 We refer to Boolean networks as systems incorporating Boolean operators or logic gates, focusing on their behavior in artificial neural networks rather than their definition in biological systems \citep{kauffman1969homeostasis}. These networks consist of interconnected nodes, each with a state represented by a Boolean variable and behavior governed by a Boolean function. Regarding the capabilities of these networks, we pose a fundamental question: 
 \begin{center} \textit{Can Boolean networks provide hypothesis functions as effective as those generated by deep neural networks?} 
 \end{center} 
 Boolean networks can exactly represent Boolean functions. Additionally, in certain contexts, these functions can be expressed through polynomial representations, even though they are fundamentally discrete and binary \citep{O’Donnell_2014}. This capability is significant as it bridges the gap between discrete Boolean logic and continuous function approximation, potentially expanding the applicability of Boolean networks across various domains \citep{NIPS2011_sumproductnet}. However, while Boolean networks are powerful in discrete domains, they do not directly conform to the Universal Approximation Theorem (UAT), which pertains to the approximation of continuous functions over real numbers \citep{hornik1989multilayer}. The UAT emphasizes the representational capacity of a model but does not address learning dynamics or generalization, which are crucial for real-world applications. 
 Given these considerations, this paper focuses on enhancing the generalization performance of Boolean networks. 
 
 The primary challenge for Boolean networks lies in their ability to generalize beyond specific, low-dimensional discrete domains. Evaluating their performance in high-dimensional and structurally complex input domains is essential. Shallow Boolean networks (bounded-depth circuits) face limitations, particularly in distinguishing certain non-uniform input distributions from the uniform distribution, which restricts their ability to efficiently capture high-order dependencies \citep{Braverman_BoundedBooleanC_2011,NEURIPS2022_5c1863f7}. Building on insights that deep neural networks' remarkable generalization performance is attributed to their hierarchical structure \citep{håstad1987computational, poggio2017and, deepshallow2017Mhaskar-Poggio}, this paper examines how Boolean networks can preserve locality and structural information through the effective organization of shallow networks in a learnable manner. Inspired by recent developments \citep{petersen2022deep}, we further adapt these methods for our specific purposes, as depicted in Figure~\ref{fig:booleannet}.

  By exploring the intriguing connections between neural networks and Boolean networks, we provide a pathway to enhance both computational efficiency and interpretability. This approach bridges theoretical foundations with practical applications and addresses crucial questions about the representational capacity and generalization abilities of Boolean networks.

\textbf{Contributions} Our key contributions in this paper are as follows:
\begin{itemize}
\item We demonstrate how deep Boolean networks can outperform traditional models by applying deep learning strategies adaptable to logic-operation-based structures.
\item We introduce a simple yet effective sampling strategy in Boolean networks to preserve spatial information in vision tasks.
\item We show that deep Boolean networks outperform Multilayer Perceptrons with reduced computational complexity and fewer parameters.
\end{itemize}

%%%%%%%%%%%%
\section{Background}\label{sec:preliminaries}

In this section, we briefly outline key preliminaries related to our Boolean networks. Boolean functions are fundamental for representing outputs from inputs using minimized formulations. For efficient implementation, hierarchical composition preserves local structures in input data, similar to decision trees. Discretized networks use predefined Boolean functions, which can be relaxed into differentiable forms, enabling backpropagation.

\textbf{Boolean Equations and Functions}
Boolean equations, first introduced by Boole in 1847, were fundamental to the development of the ``algebra of logic'' \citep{boole1847mathematical}. Shannon's 1938 work on the calculus of switching circuits further established the foundation for analyzing and manipulating logical circuits \citep{shannon1938symbolic}. Building on this groundwork, in 1973, Svoboda's Boolean Analyzer significantly advanced the computational applications of Boolean algebra, particularly in solving Boolean equations and optimizing circuits \citep{svoboda1973parallel}.

Recent research in Boolean algebra has focused on formula minimization, which aims to simplify Boolean expressions to enhance the efficiency and performance of logical circuits. A key aspect of this process is reducing logical equations to a standard form, typically $f=0$ (equivalent to $f'=1$) \citep{shannon1938symbolic}. This foundational concept enables the transformation and simplification of logical systems, facilitating effective problem-solving across various domains.

In Boolean algebra, the set of Boolean formulas on $d$ symbols $X_1, X_2, \ldots, X_d$ is denoted as $\mathbb{B}_d$. An $d$-variable function $f: \mathbb{B}_d \to \mathbb{B}$ is recognized as a Boolean function if it can be expressed by a Boolean formula \citep{brown2003boolean}. This formalization bridges Boolean formulas and functions, allowing for the representation and analysis of logical relationships in $d$-variable systems.

\textbf{Locality, Compositionality, and Hierarchy}
Boolean equations, by utilizing logical connectives, can construct increasingly complex structures. This is exemplified by the visual cortex, which has been modeled as a structural series of \texttt{AND} and \texttt{OR} layers to represent hierarchies of disjunctions of conjunctions \citep{riesenhuber1999hierarchical}. Similarly, Sum-Product Networks serve as a simple case of hierarchical modeling \citep{NIPS2011_sumproductnet}.

The concept of compositionality suggests that functions with a compositional structure can be represented with similar accuracy by both deep and shallow networks \citep{deepshallow2017Mhaskar-Poggio}. However, deep networks achieve this accuracy with fewer parameters than shallow networks~\citep{NIPS2011_sumproductnet, deepshallow2017Mhaskar-Poggio}, which lack the capacity to directly represent compositional and hierarchically local structures. For instance, while a depth-2 network with sufficient width can express all functions from $\{0,1\}^d$ to $\{0,1\}$, the network's size must increase exponentially with $d$ to achieve this~\citep{rojas2003networks, le2010deep, Livni_nips14}. The binary tree structure, with a dimensionality of 2 for two-input functions, naturally extends to hierarchical decision networks with greater depth.

\textbf{Hierarchical Decision Trees and Forests}
Decision trees form the backbone of models like random forests and define hierarchical structures for improved interpretability \citep{breiman2001random, pascha2020decisiontrees}. A decision tree $f_\theta$ is represented as $(\mathcal{G}, \theta)$, where $\mathcal{G}$ is a rooted tree structure, and $\theta = \{\theta_l\}_{l \in \textit{leaves}(\mathcal{G})}$ are parameters associated with leaves. The mapping in a decision tree is $f_\theta(X) = \sum_{l\in\textit{leaves}(\mathcal{G})}\theta_l \mathbb {I}_l(x)$, where $\mathbb {I}_l(x)$ is an indicator function \citep{khosravi2020handling}.

Decision nodes can be represented as sum units over product units, or equivalently as \texttt{OR} and \texttt{AND} units in propositional logic. Decision forests enhance predictive performance and robustness in machine learning tasks, forming a weighted additive ensemble like $F_\theta(x) = \sum_{r=1}^R w_r f_{\theta_r}(x)$ \citep{breiman1996bagging, breiman2001random}.

\textbf{Efficient Binary Operation and Relaxation of Discrete Operation}
Binary operations, including model compression and computational speed-up, significantly reduce memory usage and enhance efficiency. For instance, storing weights as bits in a 32-bit float achieves a 32$\times$ reduction in memory usage, beneficial for operations like convolutions \citep{martinez2020training, paren2022training, brown2020language}.

Boolean functions, which rely on \texttt{AND}, \texttt{OR}, and \texttt{NOT} gates, theoretically require only two logical operators (\texttt{AND} and \texttt{NOT}) to achieve completeness. In practice, however, 16 different operators ($2^{2^d}$ for $d$ variables, 16 ($=2^4$) for $d=2$) are commonly used to enhance efficiency through the integration of predefined knowledge \citep{petersen2022deep}. 
%The universal nature of \texttt{NAND} and \texttt{NOR} gates further simplifies implementations, demonstrating the effectiveness of discrete operations. 

To enable backpropagation with Boolean values $\{0, 1\}$, these values can be represented using real-valued Boolean algebra via t-norms, which are binary operations on the interval $[0, 1]$. For instance, the \texttt{AND} gate can be represented using the product triangular norm (t-norm), while the \texttt{OR} gate can be represented using the triangular conorm (t-conorm) as $A \cdot B$ and $A + B - A \cdot B$, respectively, for Boolean variables $A$ and $B$ \citep{schweizer1983probabilistic, hajek1998metamathematics}. Additional examples can be found in Table~\ref{tab:binaryoperators}. By applying this relaxation of Boolean logic to a continuous interval, logic gates are formulated with learnable parameters in neural networks~\citep{petersen2022deep}.

\textbf{Large Models and Green AI}
Large neural network models (e.g., LLMs), pre-trained on extensive datasets, serve as foundational tools for fine-tuning in downstream applications. However, this widespread fine-tuning leads to significant energy consumption and environmental impact. Moving towards \textcolor{forestgreen(web)}{Green} AI involves reducing the floating-point operations (FLOPs) in the fine-tuning process to lessen these impacts \citep{huang2023towards}.

\section{Methods}\label{sec:methods}
In this section, we explain the extension of Boolean equations and networks into deep network structures. We begin by defining the recursive form of binary Boolean equations to construct hierarchical functions, which are progressively transformed into deep Boolean networks modeled after neural network architectures. To maintain the inherent locality of spatially structured inputs (e.g. images), we introduce locality-preserving sampling and logical skip connections to deepen Boolean networks.

\begin{wrapfigure}{r}{0.60\textwidth}
\vskip -17pt
\includegraphics[width=0.60\textwidth]{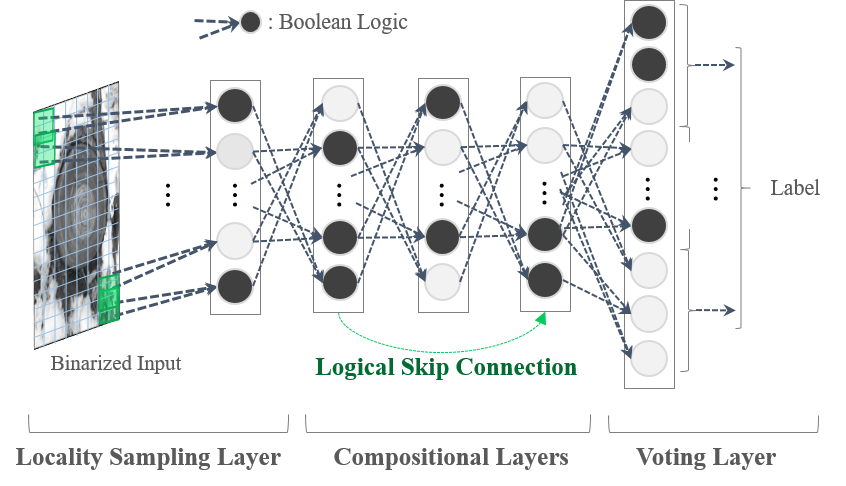}
\vskip -8pt
    \caption{Proposed Deep Boolean Networks (DBNs) include locality-preserving sampling, compositional Boolean functions, logical skip connections, and a voting layer. Further details are provided in Figures~\ref{fig:locality_sampling} and~\ref{fig:skip-connection}.}\label{fig:booleannet}
    \vskip -10pt
\end{wrapfigure}

\subsection{Deep Boolean Networks}
\begin{figure}[t]
\begin{center}
\centerline{\includegraphics[width=0.99\textwidth]{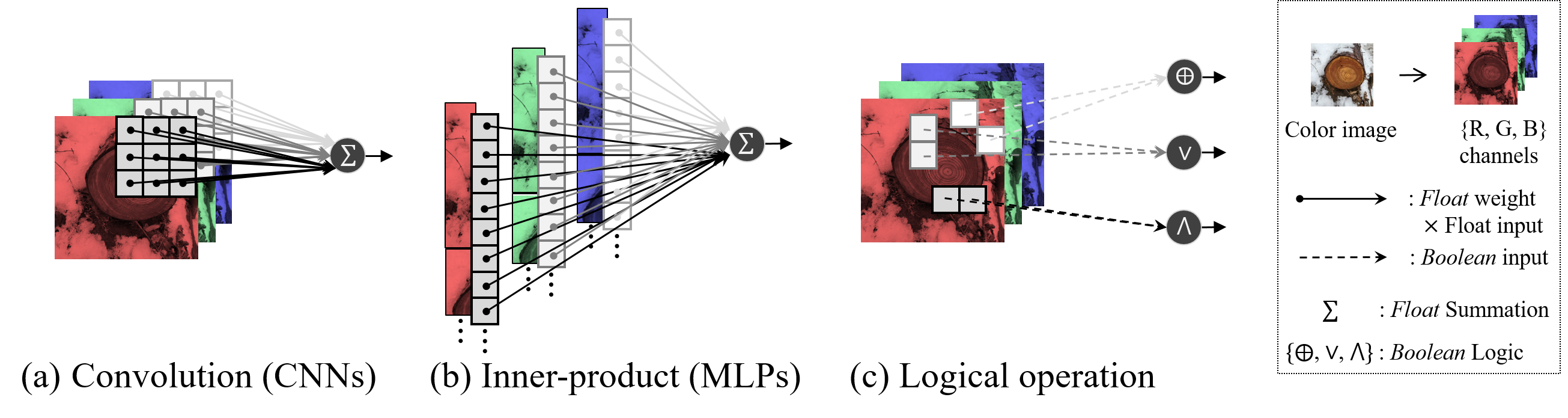}}
\vskip -0.1in
\caption{Comparison among (a) convolution, (b) inner-product, and (c) logical (boolean) operations for producing one output from an image input $\mathbb{R}^{C_{in} \times h \times w}$: (a) \textbf{Convolution} uses $C_{in} \cdot k^2$ floating-point parameters, depending on input channels $C_{in}$ and kernel size $k^2$. (b) \textbf{Inner-product} in MLPs uses $d = C_{in} \cdot h \cdot w$ parameters, based on input dimensionality. (c) \textbf{Logical (boolean) operation} requires only two or three boolean operations, in contrast to the vector multiplications in convolutions and MLPs. This is simplified to a Lookup Table (LUT) during inference.}
\label{fig:cnn-mlp-dbn}
\end{center}
\vskip -0.3in
\end{figure}

We examine deep Boolean networks, which are composed of three functional procedures: \textbf{locality-preserving sampling}, \textbf{composition of boolean functions}, and \textbf{voting-based output estimation}, as illustrated in Figure~\ref{fig:booleannet}. The baseline boolean network architecture is based on \textit{Differentiable Logic Gate Networks}~\citep{petersen2022deep}, and then modified for deeper networks. In this paper, we focus primarily on the image recognition task. Unlike convolutional operations, which are advantageous for preserving spatial information, boolean-function-based image feature extraction is not specialized in this respect. This is similar to inner-product-based operations (such as MLPs), where a flattened vector is used as input (as illustrated in Figure~\ref{fig:cnn-mlp-dbn}).

Moreover, if the 2-dimensional inputs are used for boolean functions, which are the minimum input compared to $d$-variate boolean functions, these spatial structures are rarely extracted effectively. To address this issue, we introduce the use of neighborhood pixels in the images. This approach allows very small local patches within $2 \times 2$ pixel regions to be learned following a hierarchical learning process in deep boolean networks.

A two-input function can represent the input using hierarchical composition. We define a two-input compositional function based on the work of \citep{poggio2017and, deepshallow2017Mhaskar-Poggio}:
\begin{definition}\label{def:localboolean} (\textbf{Hierarchical Composition})
Let $\mathcal{G} = (\mathcal{V}, \mathcal{E})$ be a binary tree graph, where $\mathcal{V}$ is the node set and $\mathcal{E} \subset \mathcal{V} \times \mathcal{V}$ is the edge set. For each node $i \in \mathcal{V}$:
\begin{itemize}
    \item If $i$ is a non-leaf node, then $ h_i = g_i(h_{\text{left}(i)}, h_{\text{right}(i)}) $, where $\text{left}(i)$ and $\text{right}(i)$ are the children of node $i$, and $g_i$ is a composition function associated with $i$.
    \item If $i$ is a leaf node, $h_i$ represents the function associated with node $i$.
\end{itemize}

The hierarchical composition along the binary tree is defined recursively from the leaves to the root. For leaf nodes, the function $H(\mathbf{x})$ on the input tuple $(x_m, x_{m+1}) \in \mathbb{B}^d$ is defined as:
\begin{equation}
    H(\mathbf{x}) = h_{|\mathcal{G}|} \circ \cdots \circ h_i \circ \cdots \circ h_1 \circ h_0(x_m,x_{m+1}). 
\end{equation}

For example, with an input dimension of $d=8$ and a tree depth of 3, consider the compositional function $H(x_1, x_2, \ldots, x_8) = h_2(h_{1,1}(h_{0,1}, h_{0,2}), h_{1,2}(h_{0,3}, h_{0,4}))$, where $h_{0,1}, h_{0,2}, h_{0,3},$ and $h_{0,4}$ take $(x_1,x_2), (x_3,x_4), (x_5, x_6),$ and $(x_7, x_8)$ respectively~\citep{poggio2017and, deepshallow2017Mhaskar-Poggio}. For a $d$-dimensional input, pairs of inputs are selected based on their adjacency. Using a stride of 2, this results in $d/2$ input tuples, while a stride of 1 yields $d-1$ tuples. If not constrained by adjacency, pairs are formed from any of the $\binom{d}{2}$ possible combinations.
 
To determine the compositional function at each node in Boolean networks, the optimal function is selected from a set of predefined $K$ two-input logical connectives (e.g., \texttt{AND}, \texttt{OR}, etc. See sixteen cases in Table~\ref{tab:binaryoperators} in Appendix.). By introducing coefficients, the individual contributions of Boolean functions can be adjusted. Thus, a combination of Boolean functions produces a single output, $z$, at each node through these coefficients~\citep{petersen2022deep,poggio2017and, khosravi2020handling}. The output at each node can be represented as:
\begin{equation}\label{eq:weightedsum}
    z = \sum_{j=1}^{K} {\pi}_j \cdot h_j(x_m, x_{m+1}),
\end{equation}
where ${\pi}_j$ are the coefficients corresponding to each Boolean function $h_j$ and $z$ becomes the input to the next layer. When only one coefficient is activated, it is considered the optimal function. The specific Boolean functions are detailed in Appendix, Table~\ref{tab:binaryoperators}. 

During training, these coefficients are learnable parameters.
The network uses the softmax function to transform weights ($w_j, j=1,2,\ldots,K$), which are multiplied by the gates' outputs, into a differentiable probabilistic value:
\begin{equation}
{\pi}_j = \frac{\exp(w_j)}{\sum_{k=1}^{K}\exp(w_k)} \in [0,1].    
\end{equation}  
During inference, the network selects the logic gate corresponding to the index of the most probable weight outcome. This is done by converting the probability vector $\boldsymbol{\pi}$ into a one-hot encoded vector: $\texttt{one-hot} (\boldsymbol{\pi})=[0,\ldots,0,1,0,\ldots,0 ]$ where 1 appears at the index $k$ such $k=\arg\max_j {\pi}_j$.
\end{definition}

To build on the hierarchical composition in Boolean networks, each layer applies specific transformations based on Boolean logic operations. At each layer $l$ (from 1 to $L$), a transformation or operation $\mathbf{h}^{(l)}$ is applied to the input $\mathbf{x}^{(l-1)}$ from the previous layer to produce the output $\mathbf{x}^{(l)}$, which is then passed to the next layer. This recursive operation describes how data moves forward through each layer. If $L \ge 2$, the network is referred to as a \textbf{deep} Boolean network (DBN). 

Based on the functions in Definition \ref{def:localboolean}, we introduce the deep Boolean \textit{neural} networks. The term \textit{neural} is included due to the presence of learnable coefficients that determine the Boolean logic function at each node.
\begin{definition}\label{def:Booleannet}
(\textbf{Deep Boolean (\textit{Neural}) Networks)} 
The network processes an input vector $\mathbf{x} \in \mathbb{R}^{N}$ through a binarization function $B: \mathbb{R}^N \to \mathbb{B}^{d_0}$, resulting in a binarized input $\mathbf{x}^{(0)} \in \mathbb{B}^{d_0}$. The network outputs a scalar $y_c$ for each class $c = 1, \ldots, C$, where $C$ is the total number of target classes. The network consists of $L$ hidden layers with sizes $d_1, d_2, \ldots, d_L$, each layer applying Boolean operations.

The network is parameterized by $\mathbf{W}^{(l)} \in \mathbb{R}^{d_{l-1} \times d_{l-1} \times d_l \times K}$, where $\mathbf{W}^{(l)}_{m_1, m_2, o, j}$ represents the weights associated with the input pair $(\mathbf{x}^{(l-1)}_{m_1}, \mathbf{x}^{(l-1)}_{m_2})$ for output node $o$ and Boolean function $h_j$ in layer $l$. 
Each output node $o$ in layer $l$ is computed by applying a Boolean function $h_j$ to a selected pair of inputs from the previous layer:
\begin{equation}
    z_{o}^{(l)} = \sum_{j=1}^{K} \pi^{(l)}_{m_1,m_2,o,j} \cdot h_j\left( \mathbf{x}^{(l-1)}_{m_1}, \mathbf{x}^{(l-1)}_{m_2} \right)
\end{equation}
where $\pi^{(l)}_{m_1,m_2,o,j}$ modulates the contribution of the selected input pair $(m_1, m_2)$ and the Boolean function $h_j$. Here, $m_1, m_2 \in \{1, \ldots, d_{l-1}\}$ represent the indices of the input pair from the previous layer, and the Boolean function $h_j$ is applied to the selected input pair.

For simplicity, the adjacent inputs can be selected, e.g., $m_2 = m_1 + 1$. During inference, the network simplifies by selecting only one Boolean operator per node, effectively converting the weights to binary values.

The final output for class $y_c$ is determined by the proportion of 1's in the outputs from the last layer corresponding to that class. The forward pass of the network is described as follows:
\begin{align}
\mathbf{x}^{(0)} &= B(\mathbf{x}), \\
\mathbf{x}^{(l)} &= \mathbf{h}^{(l)}(\mathbf{x}^{(l-1)}), \quad 1 \leq l \leq L, \\
y_c & = \arg\max_c \left( \frac{1}{Z} \sum_{j=1}^{d_L / C} \mathbbm{1}_{(\mathbf{x}^{(L)}_c)_j = 1} \right),    
\end{align}
    where $(\mathbf{x}^{(L)}_c)_j$ is the $j$-th element of the vector $\mathbf{x}^{(L)}_c$, which is the subset of $\mathbf{x}^{(L)}$ corresponding to the $c$-th class. The function $\mathbbm{1}_{predicate}$ is an indicator function that equals 1 if $predicate$ is true, and 0 otherwise. 

For general $n$-variable Boolean functions, the weight tensor $\mathbf{W}^{(l)}$ expands to select $n$ inputs rather than pairs:  $\mathbf{W}^{(l)} \in \mathbb{R}^{(d_{l-1})^n \times d_l \times K}$, where the first $n$ dimensions represent the selection of $n$ inputs from the previous layer. This weight tensor demonstrates the high sparsity in practice, where only $d_l \times K$ weights are active (out of the total $(d_{l-1})^n \times d_l \times K$ possible weights). This results in a density ratio of: $\frac{1}{(d_{l-1})^n}$, or during inference, $\frac{1}{(d_{l-1})^n \cdot K}$ for $\mathbf{W}^{(l)} \in \mathbb{B}^{(d_{l-1})^n \times d_l \times K}$. 

To address this sparsity, we use a more efficient implementation. In layer $l$, every node shares the same indices of nodes from the previous layer. Specifically, we use shared indices of size $d_{l-1} \times n$ for all nodes and $d_l\times K$ weights.  Compared to the full $(d_{l-1})^n$ structure in $\mathbf{W}^{(l)}$, this approach leads to more efficient memory usage while maintaining the functional capacity of the network.

%When $d_1 = d_2 = \cdots = d_L = d$, the network has uniform width $d$ and depth $L$. 

Each Boolean function in the network produces a 1-bit output, similar to a decision in a decision tree~\citep{breiman2001random}. The final classification is performed by selecting the most probable label based on the count of 1's in the output layer, implementing a voting mechanism. The normalization term $Z$ in the final layer can be learned to optimize performance, allowing for weighted voting if necessary. \end{definition}

\begin{figure}
  \mbox{}
  \begin{minipage}[t]{.48\linewidth}      \vspace{0.0pt} % Ensures the minipage starts at the top
\begin{center}
\centerline{\includegraphics[width=0.95\textwidth]{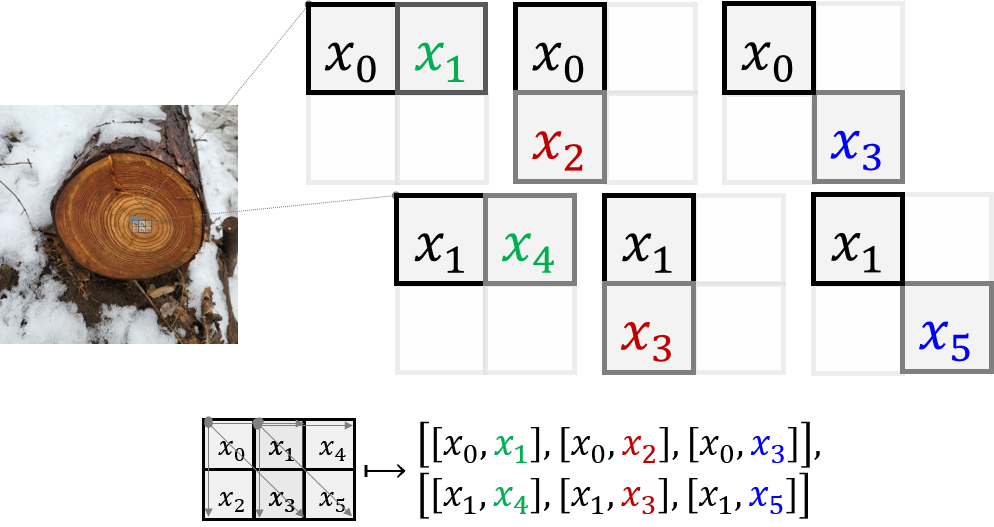}}
\caption{Examples of locality-preserving sampling in images, collecting pairwise elements from adjacent locations around a base point. Examples of pairing when the base points are $x_0$ and $x_1$ respectively.}
\label{fig:locality_sampling}
\end{center}
  \end{minipage} \hfill
  \begin{minipage}[t]{.48\linewidth}
      \vspace{-5pt} % Ensures the minipage starts at the top
   \begin{center}
  \centerline{\subfigure[MLPs]{\includegraphics[width=0.48\textwidth]{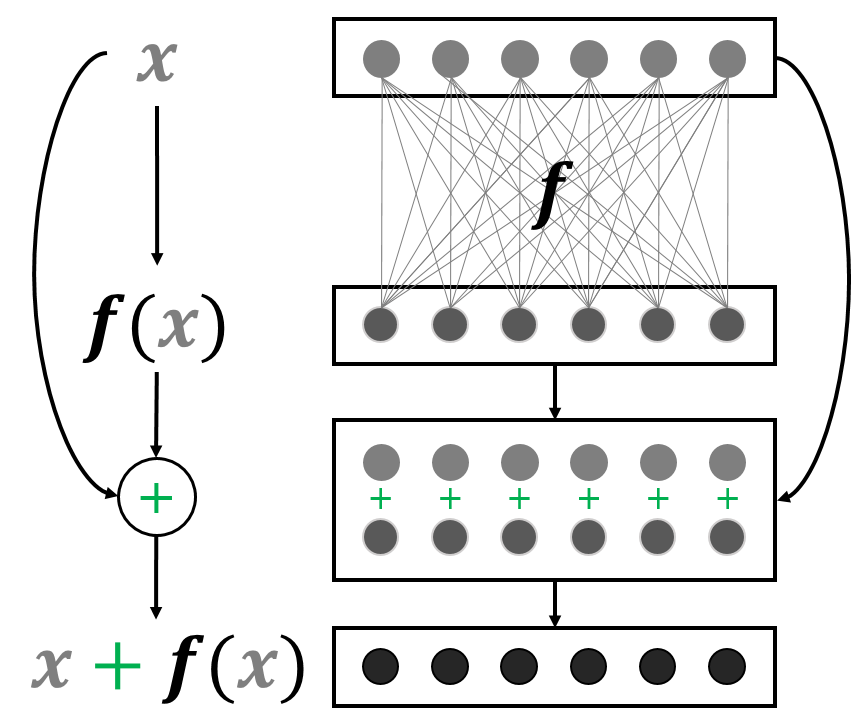}}
\subfigure[Boolean networks]{\includegraphics[width=0.48\textwidth]{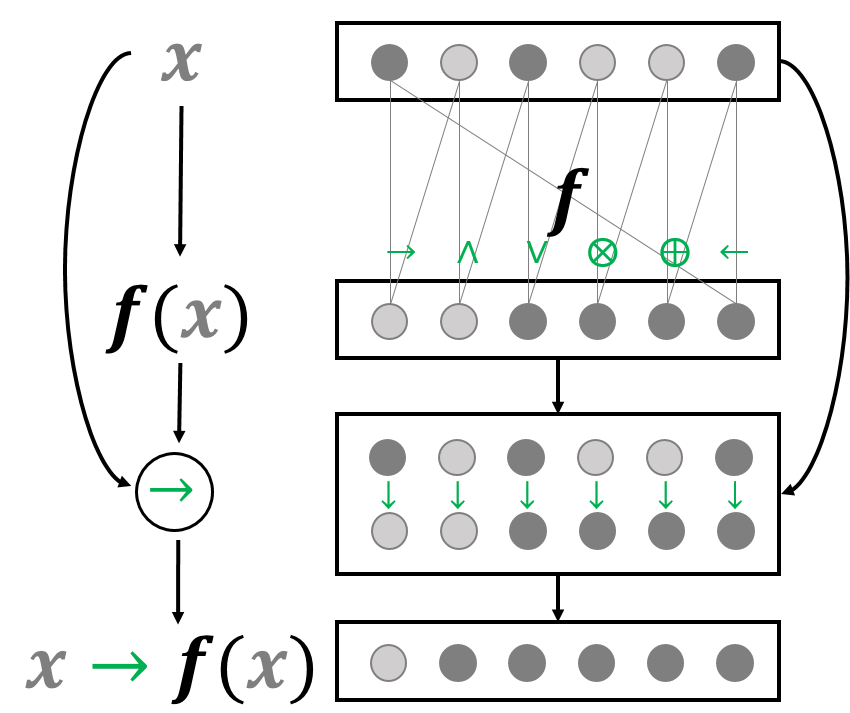}}
}
\caption{Comparison of skip connections in MLPs and Boolean networks. Unlike the fixed summation ($+$) in MLPs, boolean networks employ various logical operators can be used for the skip connection.}\label{fig:skip-connection}
\end{center}
  \end{minipage}
  \mbox{} \vskip -15pt
\end{figure}

\textbf{Spatial Locality Preserving Sampling}
To effectively capture spatial features essential for visual recognition, similar to the operation of local connections like convolutions (as illustrated in Figure~\ref{fig:cnn-mlp-dbn}), we implement a straightforward yet effective spatial input sampling strategy for networks using a vectorized input from a matrix. This approach contrasts with the global connections used in MLPs, which consider the entire input simultaneously and may lack spatial inductive bias. Due to this spatial locality-preserving mechanism, input distortions can be anticipated and effectively leveraged as a form of data augmentation. Figure~\ref{fig:locality_sampling} illustrates the locality-preserving sampling of pairwise inputs in an image. We briefly define this sampling as follows:
\begin{definition}\label{def:spatialsampling} (\textbf{Spatially Compositional Sampling Input})  Given a base point $\textcolor{blue}{x_{1,1}}$, pairwise elements are gathered from adjacent locations, forming sets such as $\{x_{1,2}, x_{2,1}, x_{2,2}\}$. In a matrix, the subscripts in $x_{row, column}$ indicates the position within the matrix. If the base point is shifted to $\textcolor{cyan}{x_{1,2}}$, the sampling would then include $\{x_{1,3}, x_{2,2}, x_{2,3}\}$. For an image of size $c \times h \times w$, this method generates $3\cdot c \cdot h \cdot w$ elements, where 3 represents sampling through three adjacent directions: row, column, and diagonal. For clarity, we provide an example illustrating a base point and its adjacent samples in each block of a matrix, highlighted with colored boxes:
\begin{tikzpicture}[baseline=(m-1-1.base)]
  \matrix (m) [matrix of math nodes, nodes in empty cells, nodes={inner sep=0pt, minimum height=10pt }   ]  {
    \textcolor{blue}{x_{1,1}} & \textcolor{cyan}{x_{1,2}} & x_{1,3}  \\
    x_{2,1} & x_{2,2} & x_{2,3}  \\
  };
   \draw[blue, thick] 
   (m-1-1.north west) -- (m-1-2.north east) -- (m-2-2.south east) -- (m-2-1.south west) -- cycle;
   \draw[cyan, thick] 
   ([xshift=-0pt, yshift=-0.5pt]m-1-2.north west) -- 
   ([xshift=0pt, yshift=-0.5pt]m-1-3.north east) -- 
([xshift=0pt, yshift=0.5pt]m-2-3.south east) -- 
([xshift=-0pt, yshift=0.5pt]m-2-2.south west) -- cycle;
\end{tikzpicture}.    
\end{definition}\vskip -11pt 
For implementation simplicity, we use columnwise and rowwise sampling, resulting in twice the image dimensionality.

\textbf{Logical Skip Connections:} To address the challenges of training deep Boolean networks, we introduce logical skip connections. These connections mitigate the vanishing gradient problem while maintaining the binary nature of the network. Figure~\ref{fig:skip-connection} illustrates the implementation of logical skip connections in our Boolean network architecture.
We define a logical skip connection layer as follows:
\begin{definition}(\textbf{Logical Skip Connection Layer}) Between two boolean inputs with the same dimensionality, a logical connection can be established element-wise using a logical operation $f$. The output of the $l$-th layer, $\mathbf{x}^{(l)}$, can be calculated with the transformed output using the transformation function of the $l$-th layer $\mathbf{h}^{(l)}(\cdot)$ for the input to the $l$-th layer $\mathbf{x}^{(l-1)}$:
\begin{equation}
\mathbf{x}^{(l)} = f(\mathbf{x}^{(l-1)}, \mathbf{h}^{(l)}(\mathbf{x}^{(l-1)})), \quad 2 \leq l \leq L-1,
\end{equation}
The function $f$ can be any logical connective, such as \texttt{AND}, \texttt{OR}, \texttt{XNOR}, or Implication Logic ((\texttt{NOT} A) \texttt{OR} B). The choice of operation allows for diverse circuit connections, enriching the expressive power of the Boolean network. Two consecutive compositional layers ($\mathbf{h}^{(l)} \circ \mathbf{h}^{(l-1)}$) to the input $\mathbf{x}^{(l-2)}$ can be applied, for example, in a bottleneck structure.
\end{definition}

Logical skip connections are integral to our hardware-software co-design strategy. Similar to traditional skip connections, which can increase memory usage by approximately 10\% in architectures like ResNets~\citep{weng2023tailor}, our logical skip connections also require additional memory to store intermediate states. However, this memory overhead could potentially be mitigated by using delay gates or other timing control mechanisms~\citep{DigitalDesign_2012}. Additionally, binary encodings require less memory than full-precision encodings, which further contributes to the efficiency of our approach. 

The use of logical operations for skip connections instead of arithmetic operations aligns well with Boolean circuit design.
Our approach differs from traditional deep neural networks by employing Boolean algebra operations instead of elementwise summation to connect hidden binary encodings. This maintains the binary nature of the network while enabling complex information flow. During the training process, the impact of logical skip connections on gradient flow varies depending on the specific logical connectives used.

\section{Experiments}\label{sec:experiments}
To empirically validate the effectiveness of our proposed method in achieving efficient generalization performance, we compare it against state-of-the-art MLP methods~\citep{bachmann2023scaling} using widely adopted benchmark datasets (CIFAR-10, CIFAR-100, and STL-10) in visual classification tasks.
\subsection{Settings}
In this paper, we conduct experiments on visual recognition tasks involving inherent spatial structures. To enable comparison with models displaying less spatial inductive bias, such as MLPs, we utilize the well-known CIFAR-10 and CIFAR-100 datasets~\citep{krizhevsky2009learning}, where both are featuring small image sizes (32 $\times$ 32 pixels), and STL-10~\citep{coates2011analysis} with larger size (64 $\times$ 64 pixels). Minimal preprocessing was applied to the images. It involves only binarization using evenly distributed 31 thresholds to convert pixel values from $[0, 255]$ to $\{0, 1\}^{31}$. For experiments with data augmentations, random horizontal flips and random scale crops are used.  

All models were trained from scratch using Adam optimizer with a learning rate of 0.01. For MLPs, we opted for SGD with a weight decay of 0.0001, which demonstrated optimal empirical results. Mini-batch size is 100 for all experiment. It's important to note that all training was conducted from scratch, without the use of pre-trained models to assess the DBN's approximation capability and generalization. All experiments were conducted on a single NVIDIA RTX A6000 GPU. Further settings are described in Appendix.

\subsection{Results}
In the context of naming model architectures (as shown in the Tables below), \textcolor{forestgreen(web)}{`DBN'} refers to a deep boolean network consisting of a sampling layer and a voting layer, as illustrated in Figure~\ref{fig:booleannet}.
\textcolor{forestgreen(web)}{`DBN-$\bullet$}' denotes a deep boolean network with a specified number of \textit{logical skip connection blocks} (\textcolor{forestgreen(web)}{$\bullet$} representing the number of blocks). For Bootleneck layer, each block with logical skip connections comprises two boolean layers. The \textit{the first layer within each block has double the output width (obtained by sampling double)} of the second layer, referring to an inverted-bottleneck structure~\citep{bachmann2023scaling}.

\textbf{Experiment 1: Comparison with the related methods.}
The baseline (DBN) consists of two layers in contrast to DiffLogicNet~\citep{petersen2022deep}, which comprises at least four layers (ranging from four to five layers with incrementally increasing widths). As our baseline, we utilized a two-layer Boolean network comprising only sampling operations and a voting layer, which does not provide the depth necessary for developing hierarchical structures. Nevertheless, by extending its depth, this baseline model was able to outperform \textsc{DiffLogicNet} on the CIFAR-10 dataset, achieving superior results with fewer parameters.
Our deeper models with fewer parameters, incorporating logical skip connections, outperform \textsc{DiffLogicNet}. \textsc{DiffLogicNet (l$\times$4)} reported a 62.14$\%$ accuracy on the CIFAR-10 dataset with 5.12M parameters. In contrast, our Deep Boolean Network (DBN) with logical skip connection variants, DBN-2 and DBN-3, achieved accuracies of 62.21$\%$ and 62.30$\%$ with only 3.23M and 3.80M parameters, respectively, as detailed in Table~\ref{tab:exp1}.
Our methods, employing deeper models with fewer parameters but without logical skip connections, demonstrated decreased performance compared to the shallower \textsc{DiffLogicNet} (5 layers) that has more parameters (5.12M), attributing this to greater width. This highlights the essential role of logical skip connections in achieving improved generalization with a reduced parameters.
Further improvements were seen with data augmentation and extended training, enhancing performance as shown in Table~\ref{tab:exp2-c10}.

\textbf{Experiment 2: Effectiveness of Data Augmentation.}
Data augmentation (DA) is a common technique used in neural networks to enhance generalization performance. This experiment aims to validate the effectiveness of DA for DBNs. By increasing the diversity of the training data, DA helps to compensate for the information loss that occurs during the binarization process. To accommodate the expanded sample space resulting from data augmentation, we extend the maximum number of training epochs to 5,000.
For comparison with MLPs, we employ the same architectures proposed in \citep{bachmann2023scaling}. In Tables~\ref{tab:exp2-c10},~\ref{tab:exp2-c100}, and~\ref{tab:exp2-stl}.

Notably, DBNs with skip connections demonstrate superior accuracy with fewer parameters on CIFAR-10/100 and STL-10 datasets. This result validates the effectiveness of data augmentation for DBNs, which outperform MLPs in accuracy while requiring significantly fewer parameters during inference.

\textbf{Experiment 3: Locality Preserving Sampling vs. Random Sampling.} We compare the difference between random sampling and locality preserving sampling. We examine randomness in all layers or the first and last layers. As shown in Table~\ref{tab:exp3-random}, more random sampling showed degraded performance compared to locality preserving sampling shown in Table~\ref{tab:exp1} and \ref{tab:exp2-c10}.

\begin{table}[t]
\caption{Classification accuracies (\%) with and without skip connections on CIFAR-10. ``\textsc{Skip}'': Skip connections; ``\textsc{BottleN}'': Bottleneck layers; ``\textsc{Rep}'': Color representation per pixel (binarized: bit $\times$ thresholds, integer: $2^\text{b} \times C$, where $b$ = bits, $C$ = channels); ``\textsc{DA}'': Data augmentation.}

\label{tab:exp1}
%\vskip 0.15in
\begin{center}
\begin{small}
\begin{sc} \scriptsize{
\begin{tabular}{lccccrcc}
\toprule
models                & no skip     & skip  & bottlen &  $\#$layers & $\#$params & Rep   & da \\
\midrule
\multicolumn{7}{l}{Baselines}\\
\midrule
DiffLogicNet (s)~\citep{petersen2022deep}   &51.27  & -     &$\times$   &    4    &48K &  1$\times$4 & $\times$\\
DiffLogicNet (m)~\citep{petersen2022deep}  &57.39  & -     &$\times$   &    4    &512K &  1$\times$4 & $\times$\\
DiffLogicNet (l)~\citep{petersen2022deep}   &60.78  & -      &$\times$   &    5    &1.28M &  1$\times$32 & $\times$\\
DiffLogicNet (l$\times$2)~\citep{petersen2022deep} &61.41  & -      &$\times$   &    5    &2.56M &  1$\times$32 & $\times$\\
DiffLogicNet (l$\times$4)~\citep{petersen2022deep} & 62.14 & -      & $\times$ &    5    & 5.12M  &  1$\times$32 & $\times$\\
\midrule
\multicolumn{8}{l}{Ours}\\
\midrule
 DBN (Baseline)             &  58.80       & -           &  $\times$ &  2 & 2.09M&$ 1\times$31 & $\times$\\
\rowcolor[gray]{.9} DBN-1            &  62.17       & 61.09       &  $\surd$    & 4  &    2.47M     & 1$\times$31 & $\times$\\
\rowcolor[gray]{.9} DBN-2            &  61.03       & 62.21       &  $\surd$    & 6  & 3.23M & 1$\times$31 & $\times$\\
\rowcolor[gray]{.9} DBN-3           &  58.92        & \textbf{62.30}   & $\surd$ & 8  & 3.80M & 1$\times$31 & $\times$\\
\rowcolor[gray]{.9} DBN-4           &  49.24         & 61.14        &  $\surd$    & 10 &  4.38M     & 1$\times$31 & $\times$\\
\midrule
\multicolumn{7}{l}{Best MLPs Baselines}  \\
\midrule
Reg.SReLUNN~\citep{Decebal2018natcom} &  68.70    & -      &  $\times$   & 3 &  20.30M  &  ${2^8}\times{3}$ & $\surd$\\
StudentNN~\citep{urban2017iclr}  &  65.80         & -        &  $\surd$   & 2 &  1.00M      & ${2^8}\times{3}$& $\surd$\\
StudentNN~\citep{urban2017iclr}  &  74.30         & -      &  $\times$  & 3 &  31.60M     & ${2^8}\times{3}$ & $\surd$ \\
\bottomrule
\end{tabular}}
\end{sc}
\end{small}
\end{center}
\vskip -0.25in
\end{table}

\begin{table}
  \mbox{}
  \begin{minipage}[t]{.48\linewidth}
 \caption{Classification accuracies (\%) for CIFAR-10 with data augmentation, averaged over 10 runs ($\pm$ standard deviation). MLPs follow \citep{bachmann2023scaling}. Trained for 5000 epochs. ``\textsc{Bottlen}'' means Bottleneck.}
\label{tab:exp2-c10}
\begin{center}
\begin{small}
\begin{sc}\scriptsize{
\begin{tabular}{llrr}
\toprule
models     &  accuracies     & params  &   bottlen \\
\midrule
\multicolumn{4}{l}{MLPs: (modules$/$width) Baselines} \\
\midrule
MLP-(6/512)       & 62.48 ($\pm$0.22)        & 2.89M  &   $\times$ \\
MLP-(12/512)      & 62.45($\pm$0.18)       &  4.47M &   $\times$ \\  
MLP-(6/1024)      & 63.83 ($\pm$0.28)       &  8.41M &   $\times$ \\
MLP-(12/1024)     & 63.71 ($\pm$0.23)       &  14.72M &   $\times$ \\
MLP-(6/2048)      & 64.74 ($\pm$0.18)        & 27.31M &   $\times$ \\
MLP-(12/2048)     & 64.54 ($\pm$0.18)        & 27.31M &   $\times$ \\
\midrule
MLP-(6/512)       & 64.17 ($\pm$0.01)      & 14.18M  &   $\surd$\\
MLP-(12/512)      & 63.99 ($\pm$0.17)          & 26.78M &   $\surd$\\
MLP-(6/1024)      & 64.54 ($\pm$0.32)          & 53.53M &   $\surd$\\
MLP-(12/1024)     & 64.33 ($\pm$0.21)          & 103.90M &   $\surd$\\
MLP-(6/2048)      & 64.64 ($\pm$0.09)        & 207.72M  &   $\surd$\\
MLP-(12/2048)     & 64.16 ($\pm$0.04)        &    409.13M  &   $\surd$\\
\midrule
 \multicolumn{4}{l}{Ours}\\
\midrule
 DBN (Baseline)     & 61.76($\pm$1.41)                        & 2.09M &   $\times$ \\
\rowcolor[gray]{.9} DBN-1     & \textbf{67.60} ($\pm$0.22)        & 2.47M &   $\surd$\\
\rowcolor[gray]{.9} DBN-2     & \textbf{67.26}($\pm$0.21)               & 3.23M  &   $\surd$ \\
\rowcolor[gray]{.9} DBN-3     & 66.53  ($\pm$0.19)                       & 3.80M &   $\surd$\\
\rowcolor[gray]{.9} DBN-4     & 65.70  ($\pm$0.16)                   &  4.38M &   $\surd$\\
\bottomrule
\end{tabular} }
\end{sc}
\end{small}
\end{center}
    \hfill 
  \end{minipage}\hfill\hfill
  \begin{minipage}[t]{.48\linewidth}
 \caption{Classification accuracies (\%) for CIFAR-100 with data augmentation, averaged over 10 runs ($\pm$ standard deviation). MLPs follow \citep{bachmann2023scaling}. Trained for 5000 epochs. ``\textsc{Bottlen}'' means Bottleneck.}\label{tab:exp2-c100}
\begin{center}
\begin{small}
\begin{sc}\scriptsize{
\begin{tabular}{llrr}
\toprule
models     &  accuracies & params & bottlen       \\
\midrule
\multicolumn{4}{l}{MLPs: (modules$/$width) Baselines} \\
\midrule
MLP-(6/512)       & 33.13 ($\pm$0.20)       &   2.94M &  $\times$  \\ 
MLP-(12/512)      & 32.75  ($\pm$0.15)        &   4.52M &   $\times$  \\
MLP-(6/1024)      & 34.86  ($\pm$0.21)        &   8.50M &   $\times$ \\
MLP-(12/1024)     & 33.99  ($\pm$0.13)        &    14.81M  &   $\times$ \\
MLP-(06/2048)     & 35.98 ($\pm$0.07)         &   27.50M &   $\times$ \\
MLP-(12/2048)     & 34.97  ($\pm$0.13)        &    14.81M  &   $\times$ \\
\midrule
MLP-(6/512)       & 37.36 ($\pm$0.23)       &    14.22M  &   $\surd$  \\
MLP-(12/512)      & 36.65 ($\pm$0.11)         &   26.83M &   $\surd$ \\
MLP-(6/1024)      & 38.62 ($\pm$0.10)        &   53.62M  &   $\surd$ \\
MLP-(12/1024)     & 38.36 ($\pm$0.13)          &   103.99M &   $\surd$\\
MLP-(6/2048)      & \textbf{39.43}  ($\pm$0.16)       &   207.91M   &   $\surd$ \\
MLP-(12/2048)     & \textbf{39.12}  ($\pm$0.18)       &    409.32M  &   $\surd$\\
\midrule
 \multicolumn{4}{l}{Ours}\\
\midrule
 DBN (Baseline)     & 35.22  ($\pm$0.30)    &  19.23M  &   $\times$ \\
\rowcolor[gray]{.9} DBN-1     & \textbf{39.46} ($\pm$0.30)    &   19.80M  &   $\surd$ \\
\rowcolor[gray]{.9} DBN-2     & \textbf{39.01}  ($\pm$0.29)  & 20.37M  &   $\surd$   \\
\rowcolor[gray]{.9} DBN-3     & 38.61  ($\pm$0.19)    &  20.94M &   $\surd$\\
\rowcolor[gray]{.9} DBN-4     & 37.75 ($\pm$0.18)    &  21.51M &   $\surd$  \\
\bottomrule
\end{tabular} }
\end{sc}
\end{small}
\end{center}
  \end{minipage}\vskip -15pt
  \mbox{}
\end{table}

\begin{table}
  \mbox{}
  \begin{minipage}[t]{.54\linewidth}
 \caption{Classification accuracies (\%) for STL-10 with data augmentation, averaged over 10 runs ($\pm$ standard deviation). MLPs follow \citep{bachmann2023scaling}. Trained for 500 epochs. Bottleneck layers are used in all cases.}
\label{tab:exp2-stl}
\begin{center}
\begin{small}
\begin{sc}\scriptsize{
\begin{tabular}{llrr}
\toprule
models     &  accuracies(std)  &  $\#$layers  & $\#$params        \\
\midrule
\multicolumn{4}{l}{MLPs: (modules$/$width) Baselines} \\
\midrule 
MLP-(6/512)     & 50.16($\pm$0.23)  & 14    & 7.87M   \\
MLP-(12/512)     & 50.57($\pm$0.29)  & 26   & 18.90M  \\
MLP-(6/1024)     & 49.71($\pm$0.19)  & 14   & 62.96M  \\
MLP-(12/1024)     & 50.06($\pm$0.24)  & 26  & 113.34M \\  
MLP-(6/2048)     & 48.88($\pm$0.29)  & 14   &  226.60M\\
MLP-(12/2048)     & 48.94($\pm$0.15)  & 26  & 428.00M \\
\midrule
\multicolumn{4}{l}{Ours}\\
\midrule
\rowcolor[gray]{.9} DBN-5     & \textbf{53.23}($\pm$0.17)  & 12 &  19.80M      \\
\rowcolor[gray]{.9} DBN-6     & 52.96($\pm$0.66) & 14 & 22.09M 
                \\
\rowcolor[gray]{.9} DBN-7     & \textbf{53.13}($\pm$0.33)   & 16 & 24.37M                    \\
\rowcolor[gray]{.9} DBN-8     & 52.45($\pm$0.33)   & 18 & 26.66M             \\
\bottomrule
\end{tabular} }
\end{sc}
\end{small}
\end{center}
  \end{minipage}\hfill  \hfill 
  \begin{minipage}[t]{.42\linewidth}
  \caption{Classification accuracies ($\%$) along random sampling through layers on CIFAR-10 dataset. Random sampling is applied for \textsc{All} layer or the \textsc{First} layer and the \textsc{Last} layer before voting layer where image locality preserving sampling is not applied.}
\label{tab:exp3-random}
%\vskip 0.15in
\begin{center}
\begin{small}
\begin{sc} \scriptsize{
\begin{tabular}{lccc}
\toprule
models     & No Skip     & \multicolumn{2}{c}{Skip} \\
\cmidrule{3-4}
           &     All        &   All    & Fist/Last layers \\
\midrule
DBN-1      &  57.15       & 56.62  & 56.10 \\
DBN-2      &  56.40       & 56.65  & 56.68 \\
DBN-3     &  53.34       & 56.83   & 56.90 \\
DBN-4     &  47.63       & 56.29   & 56.88  \\
 
\bottomrule
\end{tabular}}
\end{sc}
\end{small}
\end{center}
  \end{minipage}\hfill
  \mbox{} \vspace{-12pt}
\end{table}

\begin{table}[t]
\caption{Classification accuracies ($\%$) along different boolean logic for skip connecting on CIFAR-10 dataset.}
\label{tab:exp3}
\vskip -0.15in
\begin{center}
\begin{small}
\begin{sc}\scriptsize{
\begin{tabular}{lcccccc}
\toprule  
models      &    \scriptsize{\texttt{AND}}         &   \scriptsize{\texttt{OR}}     &    \scriptsize{\texttt{XNOR}}    & \scriptsize{\texttt{NOT} B}    &  \scriptsize{\texttt{NOT} A \texttt{OR} B }  &learned\\
\midrule
DBN-1      &   61.23    &   60.35  &  62.16  & 61.45 & 61.09 & 61.56 \\
DBN-2      &   61.45    &   61.99  &  61.47  & 62.13 & 62.21 & 61.03 \\
DBN-3      &   61.44    &   62.39  &  59.36  & 62.09 & 62.30 & 59.76 \\
DBN-4      &   62.42    &   62.04  &  57.12  & 61.95 & 61.14 & 57.01 \\
\bottomrule
\end{tabular} 
}
\end{sc}
\end{small}
\end{center}
\vskip -0.15in
\end{table}

\textbf{Experiment 4: Logical Skip Connection in DBNs.}
We explore impact if logical skip connection (refer to Figure \ref{fig:skip-connection}). The selected results are presented in Table \ref{tab:exp3}, which exhibits effective boolean logics for skip connections. Those logics (\texttt{AND}, \texttt{OR},  \texttt{XNOR}, $\texttt{NOT}~B$,  $\texttt{NOT}~A~\texttt{OR}~B$) tend to maintain better generalization as the networks deepen. Unfortunately, increasing trends are not observed in learnable skip connection logic (selected from 16 boolean logics by parameters).

From this, we introduce logical skip connections to DBNs using a boolean logic $A \Rightarrow B$, which is equivalent to the arithmetic expression $1 - A + AB$ and a combination of logic gates \texttt{NOT} $A$ \texttt{OR} $B$.

As shown in Table \ref{tab:exp1}, DBNs with \textsc{No Skip} show decreasing accuracies as the depth of DBNs deepens, while accuracies increase with \textsc{Skip}.

\textbf{Experiment 5: Deeper is better?} We evaluate effectiveness of deep layers by examining a single-layer boolean network. In the case of a single layer, it comprises the first sampling layer and the final counting layer. As shown in Table~\ref{tab:exp-single-c10} and \ref{tab:exp-single-c100}, accuracy increases with a higher width of the single layer. Notably, as expected, single-layer boolean networks do not achieve the same level of generalization as demonstrated by DBNs in Table~\ref{tab:exp1}.

\begin{table}
  \mbox{}
  \begin{minipage}[t]{.48\linewidth}
 \caption{Classification accuracies ($\%$) along different number of sampling dimensions (width) in BNs on CIFAR-10 dataset.  }
\label{tab:exp-single-c10}
\vskip -0.15in
\begin{center}
\begin{small}
\begin{sc}\scriptsize{
\begin{tabular}{lclr}
\toprule
models      & accuracies    &  width per class  & params \\
\midrule
BN      &    43.66     & w ($=19,046$) & 0.19M\\ % &  3047360 
BN      &    47.12     & w$\times2^1$  & 0.38M\\ %6094720
BN      &    50.11     & w$\times2^2$  & 0.76M\\ %12189440
BN      &    51.49     & w$\times2^3$  & 1.52M\\ %24378880
BN      &    49.27     & w$\times2^4$  & 3.04M\\ %48757760
BN      &    48.96     & w$\times2^5$  & 6.09M\\ %97515520
\bottomrule
\end{tabular} }
\end{sc}
\end{small}
\end{center}
    \hfill 
  \end{minipage}\hfill
  \begin{minipage}[t]{.48\linewidth}
     \caption{Classification accuracies ($\%$) along different number of sampling dimensions (width) in BNs on CIFAR-100 dataset.  }
\label{tab:exp-single-c100}
\vskip -0.15in
\begin{center}
\begin{small}
\begin{sc}\scriptsize{
\begin{tabular}{lclr}
\toprule
models      & accuracies    &  width per class & params  \\
\midrule
BN      &    11.56     & w ($=1,904$)     & 0.19M\\ %3046400
BN      &    13.60     & w$\times2^1$     & 0.38M\\ %6092800
BN      &    14.90     & w$\times2^2$     & 0.76M\\ %12185600
BN      &    16.54     & w$\times2^3$     & 1.52M\\ %24371200
BN      &    17.85     & w$\times2^4$     & 3.04M\\ %48742400
BN      &    22.09     & w$\times2^5$     & 6.09M  \\ % 97484800
BN      &    24.77     & w$\times2^6$     & 12.18M  \\ 
%194969600
BN      &    24.35     & w$\times2^7$     & 24.37M  \\ %389939200
\bottomrule
\end{tabular} }
\end{sc}
\end{small}
\end{center}
  \end{minipage} \vskip -10pt
  \mbox{}
\end{table}

\textbf{Experiment 6: Normalization in Voting layer.}
After last voting layer, pop-count (counting the number of 1's) is conducted. For training the networks with cross-entropy loss, the counts are normalized with a temperature value (refer to Definition \ref{def:Booleannet}), for pseudo-probability with softmax. This normalization significantly impacts generalization as shown in Table~\ref{tab:exp-teperature}, in Appendix. In our experiments, we use a value of 100 for normalization. While utilizing a learnable temperature could further enhance generalization, it doesn't seem to be the ultimate solution, leaving room for future work.

\textbf{Experiment 7: Impact of Different Discretizations of Input Images.}
For the last experiment, we investigate the impact of Cut-Out on the input image's generalization without additional pre-processing or distortion. We adjust low values in the range $\{10,20,30\}$ and high values in the range $\{225,235,245\}$ compared to the standard range $[0,255]$. Notably, the case of $[20, 255]$ demonstrates improved performance in DBN-1 as shown in Table~\ref{tab:exp-binarization}, in Appendix. We further explore this range for DBN-2 to DBN-4, with the baseline represented in the bracket using the $[0,255]$ case. Interestingly, Cut-Out appears effective for DBN-1 but not as much for deeper networks. This phenomenon may be attributed to the fact that shallower networks are more affected by these input distortions.
  
\subsection{Limitations and Future Directions}
While MLP-based Vision Transformers (ViT) do not outperform deep CNNs when trained from scratch without a large amount of data, it is common for models to not show state-of-the-art (SoTA) performance under heavy computation and pretraining constraints~\citep{vit}. Therefore, there is a necessity to embed more efficient operations.

We have not yet explored deep Boolean networks in large-scale transformer-based models. However, our proposed network uses a flat vector as an input, similar to MLPs, which should facilitate easy integration into MLP-based large models. Future work should investigate the scalability of deep Boolean networks within larger transformer architectures.

Attention mechanisms are crucial in modern large-scale models like Transformers. Integrating Boolean functions into the attention mechanism, replacing conventional multiplication operations of query, key, and value projections, offers a promising use for Deep Boolean Networks (DBNs). Our preliminary results show DBNs outperforming MLPs, consistent with the potential seen in binarized transformer networks~\citep{NEURIPS2022_5c1863f7}. The select-aggregate operation~\citep{pmlr-v139-weiss21a} mirrors attention mechanisms and suggests using Boolean operations for logical interactions between tokens. We propose applying Boolean functions to this selection-based attention method within DBNs. It has been shown that such discrete operations can be implemented using `is$\_$x' alongside MLPs~\citep{lindner2023tracr}.

%Unlike CNNs, moreover, our networks use non-shared Boolean functions, and the sampling process increases input dimensionality. This can be managed by patchifying large images and applying pooling, similar to methods in Hiera~\citep{ryali2023hiera}.

%Deploying different inputs with their corresponding truth tables allows for customized circuits tailored to specific performance criteria. This flexibility is particularly advantageous for hardware chip deployment, suggesting a promising avenue for leveraging learned Boolean networks.

\section{Conclusion} 
Our study revisited Boolean networks to enhance their generalization performance. We propose two simple yet effective strategies that prove essential for achieving good generalization: locality preserving sampling and logical skip connections. Like MLPs, Boolean networks do not inherently retain spatial information. We found that locality preserving sampling significantly improves performance by extracting and hierarchically composing useful information from images through multi-layer logical operations. Additionally, our proposed logical skip connections enable the construction of deeper Boolean networks with fewer parameters, leading to better performance. Observations showed that shallow Boolean networks, despite having many logic gates, do not achieve good generalization. Thus, the strategy to deepen Boolean networks is a crucial contribution.
Leveraging compatibility with hardware implementations, including logic gates and 1-bit Boolean algebra, we provide empirical evidence supporting the potential of Boolean networks in deep learning.  
We anticipate that our methods will inspire further investigation within the larger model community, potentially leading to new architectures that combine the strengths of Boolean networks with other deep learning approaches. Future work could explore the scalability of these techniques to larger and more complex datasets, as well as their applicability to other domains beyond image classification. 

%\clearpage

\section*{Impact Statement}
Throughout this paper, our primary focus is on the critical objective of reducing energy consumption associated with large-scale AI models. In the current landscape of the AI industry, which emphasizes the economic significance of achieving efficiency at scale, our work is positioned to make a substantial contribution to this overarching goal. We anticipate that fine-tuned versions of large models will be widely and frequently deployed globally for various downstream applications. Given their significant impact, these centralized systems should be carefully planned and implemented to mitigate potential harmful effects on our society and the environment.

Our approach advocates for the design of inherently hardware-friendly AI models, targeting the reduction of computing operations and optimization loops. By promoting such hardware-friendly AI models, particularly in the context of extensive computational platforms like large language models or generative foundational models, we believe we can significantly shift the overall environmental impact towards more sustainable and earth-friendly technology. However, those advancement in terms of hardware efficient can be always opened to misuse for enhancing harmful machines and equipment.

% \begin{ack}
% Use unnumbered first level headings for the acknowledgments. All acknowledgments
% More information about this disclosure can be found at: \url{https://neurips.cc/Conferences/2024/PaperInformation/FundingDisclosure}.

% Do {\bf not} include this section in the anonymized submission, only in the final paper. You can use the \texttt{ack} environment provided in the style file to automatically hide this section in the anonymized submission.
% \end{ack}

% \section*{References}
% References follow the acknowledgments in the camera-ready paper. Use unnumbered first-level heading for
% the references. Any choice of citation style is acceptable as long as you are
% consistent. It is permissible to reduce the font size to \verb+small+ (9 point)
% when listing the references.
% Note that the Reference section does not count towards the page limit.
% \medskip

{
\small

\bibliographystyle{plainnat}
\bibliography{logic2024}
%\bibliographystyle{abbrvnat}

%%%%%%%%%%%%%%%%%%%%%%%%%%%%%%%%%%%%%%%%%%%%%%%%%%%%%%%%%%%%

\appendix

\section{Appendix / supplemental material}
%%%%%%%%%%%%%%%%%%%%%%%%%%%%%%%%%%%%%%%%%%%%%%%%%%%%%%%%%%%%%%%%%%%%%%%%%%%%%%%
%%%%%%%%%%%%%%%%%%%%%%%%%%%%%%%%%%%%%%%%%%%%%%%%%%%%%%%%%%%%%%%%%%%%%%%%%%%%%%%
% APPENDIX
%%%%%%%%%%%%%%%%%%%%%%%%%%%%%%%%%%%%%%%%%%%%%%%%%%%%%%%%%%%%%%%%%%%%%%%%%%%%%%%
%%%%%%%%%%%%%%%%%%%%%%%%%%%%%%%%%%%%%%%%%%%%%%%%%%%%%%%%%%%%%%%%%%%%%%%%%%%%%%%
% \newpage
\subsection{Additional Experimental Results.}
Table~\ref{tab:exp-bottleneck} shows comparison between Bottleneck and Straight structures on CIFAR-10 dataset. 

Table~\ref{tab:exp-epoch-1} and Table~\ref{tab:exp-epoch-2} show the results with data augmentation on CIFAR-10 and CIFAR-100 datasets along different training epochs, respectively.

\begin{table}[ht]
\caption{Classification accuracies ($\%$) comparison between Bottleneck and Straight structures on CIFAR-10 dataset. Skip: Skip Connection, BottleN: BottleNeck Layers, B:$\#$Bits, C:$\#$Channels, DA: Data Augmentation. }
\label{tab:exp-bottleneck}
%\vskip 0.15in
\begin{center}
\begin{small}
\begin{sc} \scriptsize{
\begin{tabular}{lccccrcc}
\toprule
models                & no skip     & skip  & bottlen &  $\#$layers & $\#$params & B$\times$C   & da \\
\midrule
\multicolumn{8}{l}{Straight}\\
\midrule
DBN             &  58.80       & -           &  $\times$ &  2 & 2.09M&$ 1\times$31 & $\times$\\
DBN-1          &  60.65         &  61.07      & $\times$  & 4  && 1$\times$31 & $\times$\\
DBN-2           &  60.82       & 61.78       & $\times$  & 6  && 1$\times$31 & $\times$\\
DBN-3           &  58.21       &  61.13       & $\times$  & 8 & & 1$\times$31 & $\times$\\
DBN-4           &  52.79       & 61.34        & $\times$  & 10 && 1$\times$31 & $\times$\\
\midrule
\multicolumn{8}{l}{Bottleneck}\\
\midrule
DBN-1            &  62.17       & 61.09       &  $\surd$    & 4  &    2.47M     & 1$\times$31 & $\times$\\
DBN-2            &  61.03       & 62.21       &  $\surd$    & 6  & 3.23M & 1$\times$31 & $\times$\\
DBN-3           &  58.92        & \textbf{62.30}   & $\surd$ & 8  & 3.80M & 1$\times$31 & $\times$\\
DBN-4           &  49.24         & 61.14        &  $\surd$    & 10 &  4.38M     & 1$\times$31 & $\times$\\
\bottomrule
\end{tabular}}
\end{sc}
\end{small}
\end{center}
\vskip -0.15in
\end{table}

\begin{table}
  \mbox{}
  \begin{minipage}[ht]{.48\linewidth}
 \caption{Classification accuracies ($\%$) with data augmentation on CIFAR-10 datasets along different training epochs.}
\label{tab:exp-epoch-1}
\begin{center}
\begin{small}
\begin{sc}\scriptsize{
\begin{tabular}{llrr}
\toprule
models     &  accuracies     & params  &   epochs \\
\midrule
DBN       & 61.23 ($\pm$0.34)        & 2.09M  &   500  \\
DBN-1     & 65.22 ($\pm$0.34)       & 2.47M  &   500  \\
DBN-2     &  65.08 ($\pm$0.29)        & 3.23M   &   500  \\
DBN-3     & 64.82 ($\pm$0.26)          & 3.80M  &   500  \\
DBN-4     & 63.83 ($\pm$0.14)        & 4.38M  &   500  \\
\midrule
\midrule
DBN       & 61.76 ($\pm$1.41)                        & 2.09M &   5000 \\
DBN-1     & 67.59 ($\pm$0.22)         & 2.47M &   5000\\
DBN-2     & 67.25 ($\pm$0.21)               & 3.23M  &   5000 \\
DBN-3     & 66.53 ($\pm$0.19)                        & 3.80M &   5000\\
DBN-4     & 65.69 ($\pm$0.16)                    &  4.38M &   5000\\
\bottomrule
\end{tabular} }
\end{sc}
\end{small}
\end{center}
    \hfill 
  \end{minipage}\hfill
  \begin{minipage}[ht]{.48\linewidth}
     \caption{Classification accuracies ($\%$) with data augmentation on CIFAR-100 dataset.}\label{tab:exp-epoch-2}
\label{}
\begin{center}
\begin{small}
\begin{sc}\scriptsize{
\begin{tabular}{llrr}
\toprule
models     &  accuracies & params & epochs       \\
\midrule
\multicolumn{4}{l}{Without Data Augment}\\
\midrule
DBN      &  29.70         & 19.23M  &   500  \\
DBN-1     & 32.39       &   19.80M &   500  \\
DBN-2     &  31.96         & 20.37M  &   500\\
DBN-3     &  31.79        &   20.94M  &   500\\
DBN-4     &  31.11        &  21.51M  &   500\\
\midrule
\multicolumn{4}{l}{With Data Augment}\\
\midrule
DBN       & 35.29      &  19.23M  &   5000 \\
DBN-1     &   39.91     &   19.80M  &   5000 \\
DBN-2     & 38.93    & 20.37M  &   5000   \\
DBN-3     & 38.78      &  20.94M &   5000\\
DBN-4     & 37.90     &  21.51M &   5000  \\
\bottomrule
\end{tabular} }
\end{sc}
\end{small}
\end{center}
\end{minipage}\hfill
\mbox{}
\end{table}

\begin{table}
  \mbox{}
  \begin{minipage}[ht]{.48\linewidth}
 \caption{Classification accuracies ($\%$) along different temperatures (normalizing value) on CIFAR-10 dataset. }
\label{tab:exp-teperature}
%\vskip 0.15in
\begin{center}
\begin{small}
\begin{sc}\scriptsize{
\begin{tabular}{lcrc}
\toprule
models     &  accuracies & temperatures        \\
\midrule
DBN-1     & 52.31 &   10        \\
DBN-1     & 61.09  &   100         \\
DBN-1     & 62.34  &   1,000        \\
DBN-1     & 47.42 &   10,000         \\
%DBN-1     & 33.18  &   100000        \\
\midrule
DBN-2     &  50.33&   10    &       \\
DBN-2     &  62.21   &   100        \\
DBN-2     &  48.78  &   1,000         \\
DBN-2     & 47.42 &   10,000        \\
%DBN-2     &    &   100000      \\
\bottomrule
\end{tabular}}
\end{sc}
\end{small}
\end{center}
    \hfill 
  \end{minipage}\hfill
  \begin{minipage}[ht]{.48\linewidth}
     \caption{Classification accuracies ($\%$) along different cutout margins for image intensity on CIFAR-10 dataset.}
\label{tab:exp-binarization}
%\vskip 0.15in
\begin{center}
\begin{small}
\begin{sc}\scriptsize{
\begin{tabular}{lcr}
\toprule
models     &  accuracies & ranges            \\
\midrule
DBN-1     & 61.09  &   [0, 255]       \\
DBN-1     & 61.77  &   [0, 245]         \\
DBN-1     & 61.67  &   [0, 235]         \\
DBN-1     & 61.77  &   [0, 225]       \\
\cmidrule{2-3}
DBN-1     & 61.37  &   [10, 255]       \\
DBN-1     & \textbf{61.99}  &   [20, 255]     \\
DBN-1     & 61.55  &   [30, 255]      \\
\cmidrule{2-3}
DBN-1     & 61.61  &   [20, 245]       \\
DBN-1     & 61.87  &   [20, 235]      \\
DBN-1     & 61.49  &   [20, 225]      \\
\midrule
\midrule
%BN-1      & 59.06  &   [20, 255]        \\
%DBN-1     & \textbf{61.99}  &   [20, 255]        \\
DBN      & 58.73 (58.80) &   [20, 255]        \\
DBN-2     & 62.02 (62.21) &   [20, 255]       \\
DBN-3     & 61.82 (62.30) &   [20, 255]       \\
DBN-4     & 61.27 (61.14) &   [20, 255]      \\
%\midrule
%\midrule
%DBN-1     & 32.39  &   [0, 255]        \\
%DBN-1     &  ~31.64  &   [20, 255]        \\
%DBN-2     &  ~31.47  &   [20, 255]      \\
%DBN-3     &  31.70  &   [20, 255]       \\
%DBN-4     &  32.14  &   [20, 255]       \\
\bottomrule
\end{tabular} }
\end{sc}
\end{small}
\end{center}
  \end{minipage}\hfill
  \mbox{}
\end{table}

\subsection{Discussions}
\textbf{Comparisons to Binarized NNs:}
We attach there classification performance of the state-of-art binariy neural network, on CIFAR-10 dataset. We note that compared binary neural networks retain a 32-bit scale factor for each element which means it is not completely binarized operation. Moreover, it involves another float-precision operation, Batch Normalization (BN).
Performance significantly drops in models without BN, as explored in [Zhang et al.]. BN, while computationally expensive, especially in low-precision contexts, is crucial in training BNNs.
Additionally, we will discuss boolean approximation methods [Lowe et al.], which show that using probabilistic Boolean operations to approximate neural network activation functions can enhance performance. For instance, applying AN/OR/XNOR boolean functions improved the performance of a pretrained ResNet-18 with ReLU activation from 81.63$\%$ to 82.84$\%$.

\begin{table}[ht]
\centering
\caption{Comparison with State-of-the-Art (SOTA) binarized methods trained from scratch, excluding the use of pretrained models and incorporating data augmentation}
\begin{center}
\begin{small}
\begin{sc} \scriptsize{
\label{tab:classification_accuracies}
\begin{tabular}{llccc}
\toprule
model type & model name & accuracy (\%)& BatchNorm & SkipConnect \\
\midrule
\multicolumn{5}{c}{Binary Neural Networks (float scale factor included)} \\
\midrule
Binary CNN & XNOR-Net (ResNet-18) w/o BN [Chen et al.] & 71.75 & $\times$ & $\checkmark$ \\
\midrule
Binary CNN & XNOR-Net (ResNet-18) [Chen et al.] & 90.21 & $\checkmark$ & $\checkmark$ \\
Binary CNN & XNOR-Net (VGG-small) [Lin et al.] & 89.80 & $\checkmark$ & $\times$ \\
\midrule
Binary CNN & BNN [Hubara et al.] & 89.85 & $\checkmark$ & $\times$ \\
Binary CNN & XNOR-Net (BNN) [Rastegari et al.] & 89.83 & $\checkmark$ & $\checkmark$ \\
Binary CNN & RBNN (ResNet-18) [Lin et al.] & 92.20 & $\checkmark$ & $\checkmark$ \\
Binary CNN & RBNN (ResNet-20) [Lin et al.] & 86.50 & $\checkmark$ & $\checkmark$ \\
\midrule
\multicolumn{5}{c}{Full Precision Neural Networks} \\
\midrule
CNN (FP32) & VGG-small [Lin et al.] & 91.70 & $\checkmark$ & $\times$ \\
CNN (FP32) & ResNet-18 [Lin et al.] & 93.00 & $\checkmark$ & $\checkmark$ \\
CNN (FP32) & ResNet-20 [Lin et al.] & 91.70 & $\checkmark$ & $\checkmark$ \\
\bottomrule
\end{tabular} }
\end{sc}
\end{small}
\end{center}
\end{table}

{\footnotesize
[Zhang et al.] Hongyi Zhang et al., “Fixup Initialization: Residual Learning Without Normalization” ICLR2019

[Lowe et al.] Scott C. Lowe et al., “Logical Activation Functions: Logit-space equivalents of Probabilistic Boolean Operators” NeruIPS2022.`

[Lin et al.] M, Lin et al. "Rotated binary neural network”, NeurIPS2020.

[Chen et al.] T. Chen et al. "“BNN - BN = ?”: Training Binary Neural Networks without Batch Normalization”, CVPRW2021

[Rastegari et al.] M. Rastegari et al. “XNOR-Net: ImageNet Classification Using Binary Convolutional Neural Networks”, ECCV2016

[Hubara et al.] Itay Hubara et al. “Binarized Neural Networks”, NIPS2016
}

\textbf{Community Interest:}
While MLP-based Vision Transformers (ViT) do not outperform deep CNNs when trained from scratch without a large amount of data, it is common for models to not show state-of-the-art (SoTA) performance under heavy computation and pretraining constraints. Our methods have outperformed MLPs in image recognition tasks, though they have not exceeded the performance of CNNs. However, as ViT variants are refined, there is potential for enhancing our methods further. The most binarized versions showed degraded performance [Kim et al.], although they possess great potential in terms of computation efficiency, including binarized transformers [Liu et al.].

Importantly, Boolean networks are relatively underexplored in the context of deep learning. Given the increasing interest in neural-symbolic reasoning [Babiero et al.] and the integration of logical rules within AI models [Merril and Sabharwal], as well as recent findings by [Gidon et al.] on biological neurons computing logical gates, we believe our work opens new avenues for exploring alternative neural network paradigms. These paradigms could offer computational or energy efficiency advantages, especially in resource-constrained environments. Moreover, the concept of physics-aware training [Wright et al.] supports our argument for energy-efficient deep learning in physical systems. By opening a new venue for novel research, our work can inspire the AI/ML/DL community.

\textbf{Scope of Experiments:}
Our intention was to demonstrate the feasibility of replacing conventional modules in large models, such as projection-based operations and skip connections for multi-head attention, with our proposed Boolean network framework. Using a small network on simple image recognition tasks allowed us to validate these possibilities as a preliminary step towards adapting our methods for larger model variants. As noted in the `Limitations and Future Directions' section, scaling our approach to more complex applications will likely necessitate the development of additional regularization techniques to maintain performance and generalizability.

{\footnotesize
[Kim et al.] Minje Kim et al. “Bitwise Neural Networks”, ICML2015: proposed a bitwise version of neural networks, where all the inputs, weights, biases, hidden units, and outputs can be represented with single bits and operated on using simple bitwise logic.

[Liu et al.] Z. Liu et al. “BiT: Robustly Binarized Multi-distilled Transformer”, NeurIPS2022.

[Babiero et al.] Pietro Barbiero et al. “Interpretable Neural-Symbolic Concept Reasoning”, ICML2023: added logical rule in the prediction layer for MLP based embeddings.

[Gidon et al.] Albert Gidon et al. “Dendritic action potentials and computation in human layer 2/3 cortical neurons”, Science, 2020.

[Wright et al.] Logan G. Wright et al. “Deep physical neural networks trained with backpropagation”, Nature, 2022.
}

\textbf{Incorporation of Attention Mechanisms:}
Attention mechanisms in modern large-scale models, such as Transformers, is significant. The possibility of integrating Boolean functions within the attention mechanism—specifically replacing the conventional multiplication operations of query, key, and value projections—offers an exciting avenue for utilizing Deep Boolean Networks (DBNs). Our preliminary results, which show DBNs outperforming MLPs, are in line with the potential observed in binarized transformer networks [Liu et al.].
The `select-aggregate operation', exemplified by the RASP model [Weiss et al.], mirrors the functionality of the attention mechanism and inspires the use of discrete operations, such as Boolean operations, for logical interactions between pairs of tokens or inputs. We envision the application of Boolean logical functions to this selection operation-based attention as a novel method within DBNs. It demonstrated that such discrete operations could be implemented using `$is\_x$' alongside MLPs [Lindner et .al.]. While these methods primarily focused on interpretability, we anticipate that Boolean operations could address several limitations related to expressivity, efficiency, and realism, as described in [Lindner et al.]. Moreover, our findings reveal that formal logical rules can be utilized to describe mechanisms within Transformers, further supporting our approach [Merril and Sabharwal; Chiang et al.].

{\footnotesize
[Liu et al.] Z. Liu et al. “BiT: Robustly Binarized Multi-distilled Transformer”, NeurIPS2022.

[Weiss et al.] Gail Weiss et al., “Thinking Like Transformers”, ICML2021.

[Lindner et al.] David Lindner et al., “Tracr: Compiled Transformers as a Laboratory for Interpretability”, NeurIPS2023.

[Merril and Sabharwal] William Merril and Ashish Sabharwal, “A Logic for Expressing Log-Precision Transformers”, NeurIPS2023.

[Chiang et al.] David Chiang et al., “Tighter Bounds on the Expressivity of Transformer Encoders”, ICML2023.
}

\textbf{Deep networks with fewer parameters:}
The deeper networks can be more efficient than shallow ones, given the same number of hidden units (citing [Poggio et al.], [Montufar et al.], and [Mehrabi et al.]).

[Poggio et al.] states that ``The approximation of functions with such a specific compositional structure can be achieved with the same degree of accuracy by deep and shallow networks but the number of parameters is in general much lower for the deep networks than for the shallow network for the same approximation accuracy''

[Montufar et al.] state that ``As noted earlier, deep networks are able to identify an exponential number of input neighborhoods by mapping them to a common output of some intermediary hidden layer. … This allows the networks to compute very complex looking functions even when they are defined with relatively few parameters.''
 
{\footnotesize 
[Poggio et al.] T. Poggio et al., “Theoretical issues in deep networks”, Applied Mathematics 2020.

[Montufar et al.] G. Montufar et al., “On the Number of Linear Regions of Deep Neural Networks”, NIPS2014.

[Mehrabi et al.] M. Mehrabi et al., “Bounds on the Approximation Power of Feedforward Neural Networks”, ICML2018.
}

\subsection{Boolean equations}
\textbf{List of boolean logic:}  
In our experiment, we utilize the 16 Boolean functions defined in Table~\ref{tab:binaryoperators}. According to the design constraints required by a chip maker, specific logics can be selected and added. For example, with universal gates (\texttt{NAND} and \texttt{NOR}) and basic logic (\texttt{AND}, \texttt{OR}, and \texttt{NOT}), the deep boolean networks might need a greater depth compared to networks with those 16 logic gates.

\begin{table}[h]  
    \centering
    \caption{
        List of logic operations using two binary input values \citep{simpson1974introductory, petersen2022deep}.
    }\label{tab:binaryoperators}
    \begin{center}
\begin{small}
\begin{sc}\scriptsize{
    % \addtolength{\tabcolsep}{-3pt}
    \begin{tabular}{lccccrrr}
        \toprule
          &\multicolumn{4}{c}{$h_i(A,B)$}  & Operator                        &   Arithmetic             & Name  \\\cmidrule{2-5}
           $h_i$ $\setminus$ $(A,B)$  &   $(0,0)$     &  $(0,1)$     &  $(1,0)$     &  $(1,1)$  &                 &      &          \\
        \midrule
           $h_1$   &  0     & 0     & 0     & 0  & 0                       & $0$                 & \texttt{FALSE}\\
           $h_2$   &  0     & 0     & 0     & 1  & $A\land B$              & $AB$          & \texttt{AND}\\
           $h_3$   & 0     & 0     & 1     & 0  & $\neg(A \Rightarrow B)$ & $A-AB$              &  $A$ \texttt{AND NOT} $B$   \\
           $h_4$   & 0     & 0     & 1     & 1  & $A$                     & $A$                 &  $A$\\
           $h_5$   & 0     & 1     & 0     & 0  & $\neg(A \Leftarrow B)$  & $B-AB$              &  \texttt{NOT} $A$ \texttt{AND} $B$  \\
           $h_6$   & 0     & 1     & 0     & 1  &$B$                     & $B$                 & $B$\\
           $h_7$   & 0     & 1     & 1     & 0  &  $A \oplus B$            & $A + B - 2AB$       & \texttt{XOR} \\
           $h_8$   &  0     & 1     & 1     & 1  &   $A \lor B$              & $A + B - AB$        &  \texttt{OR}\\
           $h_9$   &  1     & 0     & 0     & 0  & $\neg(A \lor B)$        & $1 - (A + B - AB)$  &  \texttt{NOR}\\
           $h_{10}$   &   1     & 0     & 0     & 1  &  $\neg(A \oplus B)$      & $1 - (A + B - 2AB)$ & \texttt{XNOR} \\
           $h_{11}$ &  1     & 0     & 1     & 0   &  $\neg B$              & $1 - B$             & \texttt{NOT} $B$\\
           $h_{12}$ &  1     & 0     & 1     & 1   & $A \Leftarrow B$      & $1-B+AB$            &  $A$ \texttt{OR NOT} $B$ \\
           $h_{13}$ &  1     & 1     & 0     & 0   & $\neg A$              & $1-A$               & \texttt{NOT} $A$\\
           $h_{14}$ &  1     & 1     & 0     & 1   &  $A \Rightarrow B$     & $1-A+AB$            & \texttt{NOT} $A$ \texttt{OR} $B$\\
           $h_{15}$ &  1     & 1     & 1     & 0   & $\neg(A \land B)$     & $1 - AB$            &  \texttt{NAND}\\
           $h_{16}$ &  1     & 1     & 1     & 1   & 1                     & $1$                 & \texttt{TRUE} \\
        \bottomrule 
\end{tabular} }
\end{sc}
\end{small}
\end{center}
\end{table}

\textbf{Experimental packages for deep learning:}
For the experiments, we use the PyTorch framework. For more effecient dataloader for PyTorch, we use the FFCV framework.

\textbf{Network architectures:}
Here, we provide network architectures which are used in our experiments on CIFAR-10.
\begin{itemize}
    \item DBN-1:
    \begin{lstlisting}[language=Python, caption=PyToch print DBN-1 ] 
   BottleneckLN(
      (bined): Binarization()
      (flatten): BiFlatten()
      (linear_in):  BooleanLayer(190464, 190464, train)
      (linear_out): BooleanLayer(190464, 1904600, train)
      (linear_sum): VotingSum(class=10, temp=100)
      (blocks): ModuleList(
        (0): BottleneckBlock(
          (block): Sequential(
            (0): BooleanLayer(190464, 380928, train)
            (1): BooleanLayer(380928, 190464, train)
          )
        )
      )
    )
    \end{lstlisting}

    \item StandardMLP:
    \begin{lstlisting}[language=Python, caption=PyToch print MLP-12/1024 ] 
    StandardMLP(
      (linear_in): Linear(in_features=3072, out_features=1024, bias=True)
      (linear_out): Linear(in_features=1024, out_features=10, bias=True)
      (layers): ModuleList(
        (0-10): 11 x Linear(in_features=1024, out_features=1024, bias=True)
      )
      (layernorms): ModuleList(
        (0-10): 11 x LayerNorm((1024,), eps=1e-05, elementwise_affine=True)
      )
    )
\end{lstlisting}
    \item BottleneckMLP:
    \begin{lstlisting}[language=Python, caption=PyToch print MLP-12/1024 ]     
    BottleneckMLP(
      (linear_in): Linear(in_features=3072, out_features=1024, bias=True)
      (linear_out): Linear(in_features=1024, out_features=100, bias=True)
      (blocks): ModuleList(
        (0-11): 12 x BottleneckBlock(
          (block): Sequential(
            (0): Linear(in_features=1024, out_features=4096, bias=True)
            (1): GELU(approximate='none')
            (2): Linear(in_features=4096, out_features=1024, bias=True)
          )
        )
      )
      (layernorms): ModuleList(
        (0-11): 12 x LayerNorm((1024,), eps=1e-05, elementwise_affine=True)
      )
    )
    \end{lstlisting}

\end{itemize}

\end{document}